# Millimeter-wave Foresight Sensing for Safety and Resilience in Autonomous Operations

Daniel Mitchell, Jamie Blanche, Sam T. Harper, Theodore Lim, Valentin Robu, Ikuo Yamamoto and David Flynn

*Abstract*— Robotic platforms are highly programmable, scalable and versatile to complete several tasks including Inspection, Maintenance and Repair (IMR). Mobile robotics offer reduced restrictions in operating environments, resulting in greater flexibility; operation at height, dangerous areas and repetitive tasks. Cyber physical infrastructures have been identified by the UK Robotics Growth Partnership as a key enabler in how we utilize and interact with sensors and machines via the virtual and physical worlds. Cyber Physical Systems (CPS) allow for robotics and artificial intelligence to adapt and repurpose at pace, allowing for the addressment of new challenges in CPS. A challenge exists within robotics to secure an effective partnership in a wide range of areas which include shared workspaces and Beyond Visual Line of Sight (BVLOS). Robotic manipulation abilities have improved a robots accessibility via the ability to open doorways, however, challenges exist in how a robot decides if it is safe to move into a new workspace. Current sensing methods are limited to line of sight and are unable to capture data beyond doorways or walls, therefore, a robot is unable to sense if it is safe to open a door. Another limitation exists as robots are unable to detect if a human is within a shared workspace. Therefore, if a human is detected, extended safety precautions can be taken to ensure the safe autonomous operation of a robot. These challenges are represented as safety, trust and resilience, inhibiting the successful advancement of CPS. This paper evaluates the use of frequency modulated continuous wave radar sensing for human detection and through-wall detection to increase situational awareness. The results validate the use of the sensor to detect the difference between a person and infrastructure, and increased situational awareness for navigation via foresight monitoring through walls.

## I. INTRODUCTION

In 2021, there were more than 26 bn Internet of Things (IoT) devices. This number is expected to increase to 75bn by 2025 [1]. Deloitte Global's third annual survey asked 2029 companies which industry 4.0 technologies are expected to have the most profound impact on their organizations. This found IoT (72%), Artificial Intelligence (AI) (68%), cloud infrastructure (64%), big data analytics (54%) and advanced robotics (40%) as the most profound technologies for their organization [2], [3]. Radical transformation of the way products and services are offered is shaping the twenty-first century towards the necessary steps required for future automation. The amalgamation of robots, AI, advanced sensing and Digital Twins (DTs) will represent a new revolution in industry [4]. The foundations of AI, cloud infrastructure and big data analytics will be further developed by allowing the architectures and workflows of any sectors to benefit from the state of the art in technology and bidirectional comms, providing efficient decision making based on the dynamic run-time data and environment. A Cyber Physical System (CPS) is an integration of systems whose main mission is to control a physical process and, through feedback, adapt itself to new conditions in run-time. With the creation at the intersection of physical processes, networking, and computation between the physical world and the digital environment [5]. CPSs form a network for machines and assets to exchange the run-time digitalized information regarding their workload, workspace data and their health status, thus the production processes can be optimized based on the collected information.

A significant bottleneck exists in the future integration of robotics, due to industry taking a product-based approach by further developing the physical qualities such as agility, speed and balance of systems. A new perspective in the improvement of systems which enables trust and safety is paramount in the deployment of multi-robot fleets. Multi-robot fleets include several different robotic platforms, which leverage the capability of a robotic agent in an autonomous mission. With a rise in fully and semi-autonomous platforms sector wide, robotic assistants can be used to perform Beyond Visual Line of Sight (BVLOS) operations which include confined space, hazardous zones or offshore. The key motivation includes the execution of unmanned missions with lower risks and costs. In 2019, the global professional service robots market was valued at $12.3 bn and was predicted to rise by 41% in 2027 based on compound annual growth rate [6]. Considering this, in the future, automated robotic systems will require *autonomy as a service* to ensure reliability, safety, resilience and trust for a user.

Trust, safety and resilience represent the key motivations to achieve fully operational BVLOS missions. However,

*Research supported by the Offshore Robotics for the Certification of Assets (ORCA) Hub under EPSRC Project EP/R026173/1 and MicroSense Technologies Ltd in the provision of their patented microwave FMCW sensing mechanism (PCT/GB2017/053275).

D. Mitchell, J. Blanche, S. Harper and T. Lim are with the Smart Systems Group at Heriot-Watt University, Edinburgh, EH14 4AS (email: dm68@hw.ac.uk;J.Blanche@hw.ac.uk;Sam.Harper@hw.ac.uk;T.Lim@hw.ac.uk.

V. Robu was with the Centre for Mathematics and Computer Science, Intelligent and Autonomous Systems Group, CWI, 1098 XG Amsterdam, The Netherlands and Algorithmics Group, Faculty of Electrical Engineering, Mathematics and Computer Science (EEMCS), Delft University of Technology (TU Delft), 2628 XE Delft, The Netherlands (email: valentin.robu@cwi.nl).

I.Yamamoto was with Nagasaki University, Nagasaki, 852-8521, Japan (email: iyamamoto@nagasaki-u.ac.jp)

D.Flynn is with James Watt School of Engineering, University of Glasgow, UK, G12 8QQ (email: David.Flynn@Glasgow.ac.uk).

issues lie in the quantification of trust, safety and resilience; how can we increase transparency in a system? For fully autonomous missions, a robotic platform must be required to evaluate its own state of health, risk and likelihood of mission success. The integration of these systems must be paired alongside digital versions of themselves and the environment they are deployed in to maximize reliability and mission success. Providing this hyper-enabled overview builds trust and allows remote operators to clearly identify, "is my robot safe", "are there any faults within my robotic platform?" and "are there any personnel in a shared workspace with a robot?".

This paper presents the development of foresight sensing for a cyber physical symbiotic system, which integrates novel Frequency Modulated Continuous Wave (FMCW) sensing to detect the presence of humans though walls or doors. Extending the sensing capabilities of a RAS out with the immediate mission space demonstrates the importance of trust and resilience via an increase in situational awareness. Allowing for real-time decisions to be made for examples such as "Is there a human on the other side of a doorway?" and "Is there enough space for a robot to access beyond the initial doorway?" To evaluate a low-power and non-contact sensing solution, an FMCW radar sensor was installed as a payload onboard a Husky A200 with dual UR5 manipulators.

This publication is structured as follows. In Section II, the current state of technology is discussed for through-the-wall sensing methods and robotic safety features. Section III discusses the importance or trust, safety and resilience for autonomous systems. A new methodology, foresight sensing, for sensing requirements of autonomous systems is highlighted in Section IV. The results are presented in Section V, where human proximity alerting and through-wall detection is validated. Section VI presents the discussion, followed by the conclusion in Section VII.

## II. CURRENT STATE OF TECHNOLOGY

This section identifies how state of the art sensing methods are used for through-wall detection and how other safety features are being implemented onboard robotic platforms.

### A. Through-wall Detection

Sun *et al.* investigated the use of a through-the-wall sensing method based on forward scattering for detection of humans. It is demonstrated that the transceiver radar is capable of detecting various types of human motions behind the wall with positive sensitivity and at very low transmission powers [7]. However, this method required the use of two antennas (Tx and Rx) to be positioned behind each wall, detracting from field deployability.

Ma *et al.* report a corner multipath for through-wall radar imaging based on compressive sensing. The sensing mechanism utilizes a linear array aperture of several wideband transceivers placed parallel to a wall, which can detect the positions of objects around the corner of the wall [8]. Limitations of this sensing method include the requirement for several antennas and the inability of detect signal contrasts for differing material types, aiding target identification.

Alkus *et al.* utilize a W-Band millimeter-wave FMCW radar which utilized compressive sensing to illuminate a wall and target on the opposite side of the wall. The sensor accurately localized a 12 mm, metallic object behind 3mm thick drywall, however, the graphical results require improvements due to low output image resolution, resulting in difficulties in target identification [9]. The sensor must also be tested against thicker and more dense wall types.

Yanik *et al.* discusses the use of FMCW radar sensing in the W-band for the detection of concealed item detection. This system has very accurate detection of the shape of concealed items, however requires a large setup including X and Y Axis motors to construct the resulting image. The sensor has also not yet been tested for the detection of items behind more dense materials [10].

A summary of the key properties of the previously discussed sensing methods is discussed in Table I. The table identifies that many of the methods only require a single sensor and can detect the presence of different objects through-the-wall. However, a limitation exists as these sensing methods are unable to detect and distinguish the difference in the presence of a human versus an object. If a sensor was able to do this, then they would be able to update safety protocols for the robot ahead of entering a new workspace.

### B. Robotic Safety Features

A methodology to increase safety of mobile robotic platforms is proposed within [11], which proposes a method to overcome the issues in human-robot collaboration which exist in caged robots and robots which operate behind light curtains. To implement the methodology, a sensing suite is required to enable safe operation of the robot where a tiered hierarchy which represents layered zones around the robots exist. An additional feature includes privileged access to certain well-trained individuals.

A real-time human-robot collision safety evaluation method for collaborative robots (cobots) is proposed within [12]. The paper discusses the use of a run-time algorithm and mathematical model with the ability to estimate the pressure associated with different tools and materials upon a collision. Pig skin was used within the investigation due to similar compression properties with the soft skin of a human forehead. The algorithm is effective at reducing the estimated contact force if a collision does occur however, it does seem that the algorithm simply slows down the speed of movement of the manipulator arm to allow the force to be below the predetermined 'allowable force'. Limitations exist within this approach which limit the productivity of the cobot. This approach could be improved by inclusion of a sensor which can detect if a robot is in close proximity to allow the algorithm to be switched on to reduce the risk of a collision by slowing down the cobot to the predetermined 'allowable force'.

TABLE I  KEY PROPERTIES OF THE DISCUSSED SENSORS IN SECTION IIII.A.

| Reference | Single sensor | Object Detection | Distinguishing of Human |
|---|---|---|---|
| [7] | ✔ | ✔ | ✗ |
| [8] | ✗ | ✔ | ✗ |
| [9] | ✔ | ✔ | ✗ |
| [10] | ✔ | ✔ | ✗ |

An interactive human-robot collision avoidance system approach was implemented for mobile robots which are deployed in shared environments indoor. The system utilized a Kinetic 2.0 sensor as a payload onboard a H20 robot. When a human is in the path of the robot, the robot firstly tries to interact with the human to execute several actions based on arm positions from a human; collision avoidance, move forward, move backwards, left or right. The human must use specific gestures to interact with the robot [13]. If the robot receives no gestures, then collision avoidance occurs. Limitations exist in the scenario where a human is holding objects, therefore unable to complete the required gestures, or on the occasion where someone is disabled, in addition to the scenario includes someone who is untrained and is in the path of the robot. Another limitation occurs due to the robot relying on visual line of sight with a human. The robot may find difficulties at corners of hallways or at doorways.

In summary, several of the previously discussed safety methodologies have several advantages and disadvantages. Safety in design has been critical in establishing these safety features for robotic and autonomous systems. However, key limitations exist due to the following results displayed in Table II. In the case studies utilized, many of the methodologies utilized tiered levels of safety. This is identified as an advantage for robotics as it maintains the productivity of robots. In many cases, when a human is in a shared workspace, robotic speeds are limited to ensure a human can predict a robotic motion or evade a robot in the event of a potential collision. The detection of objects in a workspace includes a robots ability to detect an object and avoid the object. Therefore, [12] is excluded as the algorithm can only detect a collision during the incident. This results in a limitation as the algorithm relies on a breach of safety. This should be implemented as a last resort hence a barrier in the system. Autonomous systems must have the ability to detect and distinguish humans that enter a workspace. The methodology proposed in [11] has the ability to detect a human if they have the required access card, however, it is not clear if the sensing suite has the ability to detect a human beyond collision avoidance if a human enters a workspace without an access card. Due to the dynamic environment and purpose of several robots, they will often be deployed in hazardous environments, therefore, robots must have the ability to operate safely in opaque environments. None of the papers reviewed discussed this scenario, however, should be considered in the future where there may be unforeseen circumstances.

### III. TRUST, SAFETY AND RESILIENCE FOR ROBOTIC PLATFORMS

Advances in RAS mobility have led to the deployment of autonomous systems in more advanced/difficult environments including the offshore, nuclear, manufacturing, space and public sectors. This has created several opportunities for CPSs in parallel to several barriers which inhibit the successful deployment of robotics. These relate to trust, resilience and safety issues (as viewed in Fig. 1) where robots are expected to consistently operate effectively. Resilience includes the ability for an autonomous system to recover from, or adjust to, misfortune or change. From the perspective of robustness, safety compliance and reliability: resilience means an autonomous system can adapt its planned mission strategy by operating key systems irrespective of adversity. This reduces the risk of an unsuccessful mission. Resilience relates to robotics as a resilience framework prevents the loss of assets, minimizes costs and ensures recovery from high impact/low frequency events. An example of this includes where primary navigation sensors are compromised due to low visibility conditions, such as fog, mist, smoke, and where supporting navigation sensing (FMCW radar sensing) is used to return to a safe position.

Trust is an important relationship which must be gained by the human operators to ensure safe deployment of robotics and artificial intelligence. This is especially important in shared, collaborative workspaces, where humans must work in shared environments with robots. Some companies which utilize robotics do not have shared workspaces, resulting in robots that are often caged off (to mitigate risks) from humans in nearby proximity. This is due to risks that exist as minor robotic mistakes can lead to significant safety risks. However, more applications of service robots are implemented in areas, such as warehouses, as environmental conditions are well lit and clear of obstacles, reducing the risks of collisions. This allows for robots to avoid collisions and operate within shared workspaces.

TABLE II   SUMMARY OF THE KEY PROPERTIES OF THE SAFETY FEATURES DISCUSSED IN SECTION IIII.B.

| Reference | Tiered levels of safety features | Detection of objects in a workspace | Detection of humans | Operational in opaque conditions |
|---|---|---|---|---|
| [11] | ✓ | ✓ | ✓/✗ | ✗ |
| [12] | ✗ | ✗ | ✗ | ✓ |
| [13] | ✓ | ✓ | ✓ | ✗ |

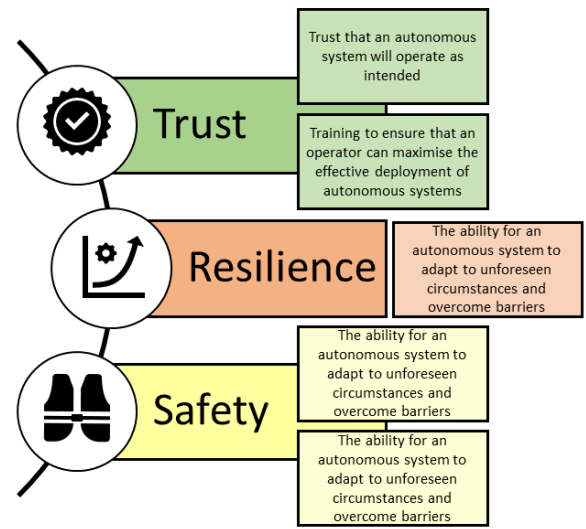

Figure 1. Key barriers in the deployment of fully autonomous systems and descriptions of challenge examples.

Robotic safety standards are a collection of guidelines for robot specifications and autonomous operations in which all involved in the manufacture, sales and use of robots must adhere to. Often a three-stage approach is utilized when manufacturing and designing safety standards. This includes safe design, protective measures and warnings and instructions. Safety standards do exist for robotics and autonomous systems including EN ISO 12100 (Safety of machinery), EN ISO 10218 (Safety requirements for industrial robots) and EN ISO 13482:2014 (Safety requirements for personal care robots) [14]. However, due to the rapid advancement of robotics, it can be difficult to ensure safety standards are maintained and updated alongside the advancement of robotics and artificial intelligence [15].

### IV. FORESIGHT SENSING

An area where minor advancements can create significant improvements in trust and safety includes advanced sensing mechanisms. As identified within this literature review there exists a gap for a compact, lightweight, low power sensing mechanism with the ability to sense through walls to allow for an increased operational overview.

Foresight sensing allows for a robotic platform to have increased situational awareness via the ability to sense the difference between an object, a person and infrastructure. This also has an added capability of through-wall detection of objects in close proximity to doors and walls. This leads to the following benefits for autonomous systems:

1. Increased safety precautions due to a robot having the ability to detect a person.
2. Increasing navigational and situational awareness for autonomous operation between workspaces.

The Clearpath Husky A200 robot (Fig. 2) is an autonomous ground vehicle with a pair of dual UR5 manipulators. The platform ha payloads of a 3D Velodyne LiDAR alongside a 2D SICK Lidar positioned on the front of the base. A pan-tilt unit was utilized for effective positioning of the FMCW radar sensor onboard the robotic platform. In previous autonomous missions, the FMCW radar was utilized for the following as an asset integrity role:

1. Detection of surface corrosion precursors on metals [16], [17].
2. Monitoring of surface and subsurface defect precursors in wind turbine blades [18].
3. Analysis of civil infrastructure for subsurface fault precursors [19].

However, for this investigation, the radar was utilized to support a new use case under the term foresight sensing. The millimeter wave sensing unit can be used to support navigational sensors in scenarios which reduce the reliability of conventional sensing techniques such as smoke, steam and mist. This allows for resilient sensing in dynamic offshore environments and scenarios.

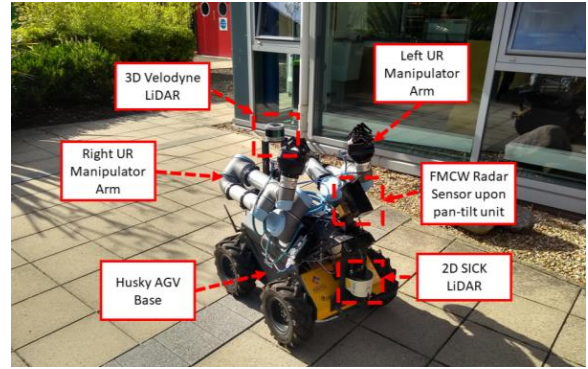

Figure 2. Dual UR Husky A200 AGV fully annotated highlighting sensing payloads and key components.

Foresight sensing feeds into cyber physical systems due to the benefits of an increased operational overview for autonomous systems and increased symbiosis between field deployed humans and robots. This allows for increased capability as a robot can update its safety features in the presence of a human.

### V. RESULTS

Robotic resilience is a primary aspect which must be considered for the deployment of an asset. Mobile autonomous platforms are designed to be deployed in hazardous environments such as opaque conditions. These include areas of steam, mist and smoke. These conditions are commonly found in the offshore renewable energy sector, oil and gas sector, mining and search and rescue operations. Autonomous robotic platforms must have supporting navigational tools which can ensure the resilience of a mission or support an '*Adapt and Survive*' methodology which will ensure the integrity of the asset.

The FMCW radar sensor is an emergent and critical asset which will ensure that a robotic platform can remain resilient in testing environmental conditions. The following discusses how the FMCW radar can be applied for the detection and distinguishing of humans within a shared workspace and for through-wall detection to ensure safety compliance of robotic platforms.

#### A. Human Proximity Alerting

The following application of the FMCW radar sensor investigates human proximity alerting. Millimeter-wave sensing enables distinguishing of a human and an asset or structure by the autonomous platform. This improves the situational awareness of a platform, ensuring it meets safety compliance requirements in terms of robot proximity to humans and infrastructure, representing a major function of SLAM. This application summarizes the use of the FMCW sensor to distinguish a human target and a metallic target at a range of distances from the robotic platform.

The FMCW sensor and robotic platform remained in a static position where the human represented the variable as they repositioned themselves further from the sensor. The human was positioned directly within the Field Of View (FOV) of the FMCW sensor and moved in 1-meter increments to 4 meters away from the sensor. This methodology was then repeated where the positioning of the human was replaced with an aluminum metal sheet measuring 700mm by 500mm. The Return Signal Amplitude (RSA) responses as a function of distance from the radar sensor can be viewed within Fig. 3. The figure has been annotated to aid with the visualization of the results presented. The diagram firstly presents an empty reference signal for when there is no target in the FOV of the radar sensor. The RSA identifies the back wall of the laboratory as annotated with the blue block at 6.5m. The diagram utilizes solid lines with a yellow block for when the Aluminum metal sheet was positioned in the FOV of the radar sensor and dashed lines alongside a human silhouette for when the human was positioned in the FOV. The data collected within this investigation presents a difference in RSA where an algorithm can utilize the peaks collected and produce a reflection magnitude relative to a set baseline, in this case the laboratory wall at 6 meters from the sensor as shown in Fig. 3 [20]. It must be noted that the FMCW radar was not properly calibrated for the investigation, therefore the distances to the target are slightly increased within Fig. 3.

Equation (1) was used to calculate the relative reflection magnitudes by utilizing the RSA for the empty reference against the RSA for the target as displayed in Table III.

$$RRM = \frac{Reference\ RSA}{Target\ RSA} \qquad (1)$$

Where Relative Reflection Magnitude = RRM.

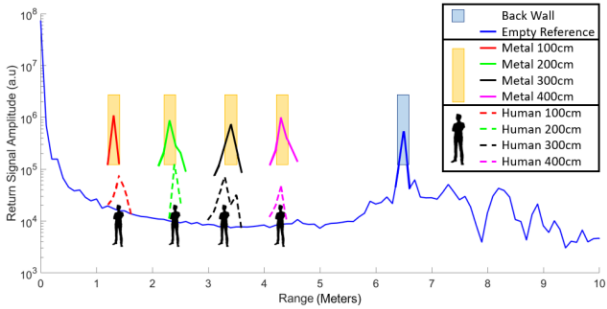

Figure 3. RSA versus distance for human and metal sheet targets.

TABLE III   RELATIVE REFLECTION MAGNITUDES CALCULATED FROM THE FMCW RADAR SENSOR DATA

| Target Type and Distance from Sensor | Relative Reflection Magnitude |
|---|---|
| "*No Target*" Reference (or Lab Wall at 600 cm) | 1 |
| Human at 100cm | 1.55 |
| Human at 200cm | 1.88 |
| Human at 300cm | 1.51 |
| Human at 400cm | 1.32 |
| Aluminium Sheet at 100cm | 14.93 |
| Aluminium Sheet at 200cm | 10.79 |
| Aluminium Sheet at 300cm | 7.51 |
| Aluminium Sheet at 400cm | 13.52 |

## B. Through-wall Detection

The following use case demonstrates the capability of FMCW for detection of objects through walls. This initial work was conducted to assess whether the FMCW radar could be used to increase the situational awareness of a robotic platform to variables, such as a human or robotic platform entering a room or mission space. This level of situational mapping promotes the safety compliance of autonomous systems when operating in constricted, high foot-traffic areas. For example, if a human is working in close proximity to a regularly used door, if the FMCW equipped robot can detect the human as an obstacle obscured by the door, then the autonomous platform can replan to find entry via another route. This would increase safety and trust of robotic platform by mitigating potential human collision risks during transit through the doorway or by avoiding becoming stuck between the door, rendering it unusable. This risk is increasing as several platforms such as SPOT, Robotnik and Dual UR5 Husky have manipulator arms with the ability to access and open doorways.

Fig. 4 illustrates where the radar was positioned 10 cm from the FOV of a partition wall within the laboratory. A copper sheet measuring 30 x 30cm was held by a human in the hallway on the other side of the wall from the radar. The sheet was moved from wall 2 towards wall 1 whilst the radar was continuously scanning. Fig. 5 shows the results, where an empty reference baseline was established for the partition wall (wall 1) and hallway to wall 2. The human holding the copper sheet was then positioned with their back against wall 2. The copper sheet was therefore at position A and moved continuously towards the radar until the copper sheet was pressed against wall 1. Fig. 5 displays observed peaks within the RSA from the empty reference baseline indicating where the copper sheet is detected by the radar. The red, green, black and cyan peaks indicate positions A-D respectively and demonstrate detection of the copper sheet when moving towards the radar whilst obscured through the wall.

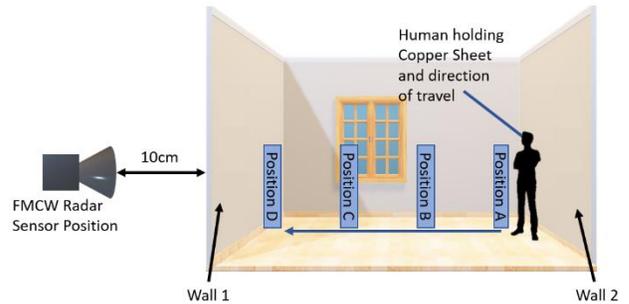

Figure 4. Diagram highlighting the procedure of the through wall detection investigation.

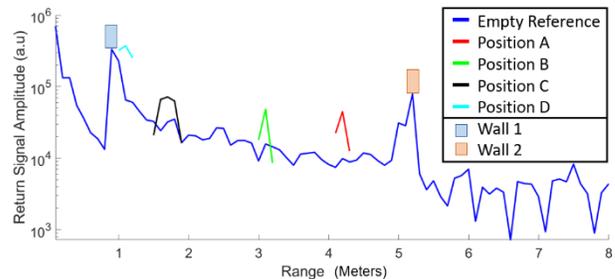

Figure 5. RSA responses for through wall detection investigation at positions A-D.

## VI. Discussion

This paper discussed the utilization of millimeter-wave radar sensing for human detection and through wall detection of obstacles (in this case a human holding a copper sheet) on the opposite side of a partition wall. The FMCW radar sensor represents a novel payload which could be used by autonomous systems for the detection of humans in line-of-sight scenarios. For example, if a human entered a workspace. This would allow for a robot to automatically update safety standards if a human entered a collaborative workspace.

Due to the resilient sensing properties of FMCW radar, the sensor could also be utilized to support primary navigational sensors in the case of opaque conditions (smoke, steam or mist). These conditions often reduce the accuracy of visual spectrum and laser based navigational systems such as cameras and LiDARs. Therefore, FMCW radar sensing can be utilized to support navigation in these unforeseen circumstances to ensure mission resilience and safety compliance.

Through-wall sensing is an emerging sensing technique which will be important in ensuring trust and safety compliance of autonomous systems in the future. This is due to the sensor having the ability to update path planning and/or update safety rules ahead of entering a workspace. This increases situational awareness for an autonomous system to ensure that it operates within predetermined safety regulations.

Search and Rescue environments are often considered highly variable and harsh conditions, with low visibility, thermal extremes, confined spaces, infrastructural failure and noxious fumes all acting as contributing factors for human disorientation or incapacitation. In this paper, we demonstrated the ability of the FMCW system to detect static and dynamic features through partition walls and in other publications utilized the sensor to assess more dense materials such as sandstone and concrete [19], [21]. This leads to the potential unique use of FMCW for person and object detection in the built environment. The FMCW system demonstrated within this paper is capable of functioning in conditions which inhibit more standard methods of robotic platform localization, such as LiDAR and visual cameras.

Some applications where FMCW radar may be developed for use in search and rescue are as follows:

1. Discrimination between humans and infrastructure, allowing for localization of vital rescue priorities in areas where normal communications have failed, such as GPS-denied environments.
2. Detection and identification of humans within a search and rescue area, even when obscured by failed infrastructure or in low visibility conditions.
3. Detection of key vital sign data for assessment of casualties [22]–[25].
4. Support navigation and mapping in opaque conditions to provide emergency service commanders and personnel with run-time situation reporting/mapping.

A key barrier which inhibits the advancement of robotics and AI includes trust. This requires a balance between consistency and capability to ensure trust between human operator and robot. This is important in terms of the resilience and safety of a mission if a robot is deployed BVLOS however, is increasingly important for robots in shared, collaborative workspaces due to increased risks for humans in shared spaces. A loss of trust due to an accident can be extremely detrimental to the deployment of robots regardless of the benefits in productivity as safety is the most imperative feature of any autonomous system. Therefore, the FMCW radar and other sensing mechanisms will ensure that a human is effectively detected and AI can enable for tiered levels of safety to be enabled.

In the roadmap to advancing autonomous systems, robots will require increased sensing mechanisms and algorithms as discussed in literature this literature review and novel work presented in this paper to increase situational awareness, trust, safety and resilience. However, the authors of this paper view a Symbiotic System of Systems Approach (SSOSA) as a key enabler to unlocking the future potential as robots move from single use cases to multi-robot fleets [26]. This can be enhanced via bidirectional communications across a cyber physical systems infrastructure which integrates and synchronizes models, data and information streams from deployed robots. This includes sensing, actuators and run-time data via bidirectional communication. To support trusted deployments, information is fed to a synthetic environment (DT) to ensure the orchestration and control of robots, missions and objectives alongside the analyzation of data, results from IMR and self-certification data from robots. This demonstrates the importance of this approach of future integrated autonomous systems as the real and digital world blend seamlessly together for efficient utilization of information. To ensure the resilient, safe and trusted deployment, *'Autonomy as a Service'* will also include a predominant role for functions including operation, maintenance and optimization within infrastructures.

## VII. Conclusion

Safety, resilience and trust are key components in a successful autonomous mission. It is clear that a variety of sensing mechanisms and algorithms will be required to advance these factors for robotic and autonomous systems. The application of this research offers a solution to effective detection and distinction of humans. In addition to increased situational awareness to update navigational and safety parameters via through wall detection using the FMCW radar sensor. Foresight sensing allows for robotic platforms to have increased situational awareness for BVLOS sensing to ensure safety standards are met whilst overcoming barriers to mission success. To increase trust in robotic agents, systems must operate first time, every time. This will lead engineers to better understand and trust the productivity created via a robot. This will enhance cyber physical systems where robots can become more symbiotic via encouraging effective partnerships across environments, infrastructure and personnel, leading to cyber physical symbiotic systems.

ACKNOWLEDGMENT

The research within this paper has been supported by the Offshore Robotics for Certification of Assets (ORCA) Hub [EP/R026173/1] and the authors wish to thank MicroSense Technologies Ltd for use of their patented FMCW radar sensor (PCT/GB2017/053275).